\documentclass[runningheads]{llncs}
\usepackage{graphicx}
\usepackage{tikz}
\usepackage{comment}
\usepackage{amsmath,amssymb} % define this before the line numbering.
\usepackage{color}
\usepackage{vcell}
\usepackage[width=122mm,left=12mm,paperwidth=146mm,height=193mm,top=12mm,paperheight=217mm]{geometry}

\usepackage{hyperref}

\makeatletter
\newcommand{\printfnsymbol}[1]{%
  \textsuperscript{\@fnsymbol{#1}}%
}
\makeatother

\begin{document}
% \linenumbers
\pagestyle{headings}
\mainmatter
\def\ECCVSubNumber{1}  % Insert your submission number here

\title{``Knights'': First Place Submission for VIPriors21 Action Recognition Challenge at ICCV 2021} % Replace with your title
% \begin{comment}
% \titlerunning{ECCV-20 submission ID \ECCVSubNumber} 
% \authorrunning{ECCV-20 submission ID \ECCVSubNumber} 
% % \author{Anonymous ECCV submission}
% \institute{Paper ID \ECCVSubNumber}
% \end{comment}
%******************

% \begin{comment}
% \titlerunning{``Kallis'}
\author{Ishan Dave\inst{1}\thanks{equal contribution}, Naman Biyani\inst{2}\printfnsymbol{1}, Brandon Clark\inst{1}, Rohit Gupta\inst{1}, \\ Yogesh Rawat\inst{1} \and Mubarak Shah\inst{1}}

\authorrunning{Dave et al.}
\institute{Center for Research in Computer Vision (CRCV), University of Central Florida, Orlando, Florida, USA \and
Indian Institute of Technology, Kanpur, India, \\
\email{\{ishandave, brandonclark314, rohitg\}@knights.ucf.edu,\\
namanb@iitk.ac.in \{yogesh, shah\}@crcv.ucf.edu}
}

% \end{comment}
%******************
\maketitle
% \footnote{* denotes euqal contribution}
\begin{abstract}
% \footnote{* denotes euqal contribution}
% This paper presents our approach ``Kallis'' to solve the action recognition task on UCF101 without using any pre-trained weights. We proposed a two-stream architecture using various spatio-temporal resolutions that aid in learning the long and short-range temporal structures of activities at different spatial scales. We also show that taking an average of the predictions from multiple clips that have various spatio-temporal resolutions and augmentations helps increase the performance as well as significantly lowers the training cost. Without using any pre-trained weights, the proposed solution achieves a Top-1 accuracy of \textbf{90.83\%} on UCF-101 (split 1) test set and a Top-1 accuracy of \textbf{90.71\%} on the test set of the Visual Inductive Priors for Data-Efficient Deep Learning Workshop’s Action Recognition Challenge, ECCV 2020, which is the best among all of the other entries.  

This technical report presents our approach \textit{``Knights''} to solve the action recognition task on a small subset of Kinetics-400 i.e. \textit{Kinetics400ViPriors} without using any extra-data. Our approach has 3 main components: state-of-the-art self-supervised pretraining, video transformer models, and optical flow modality. Along with the use of standard test-time augmentation, our proposed solution achieves \textbf{73\%} on Kinetics400ViPriors test set, which is the best among all of the other entries \textit{Visual Inductive Priors for Data-Efficient Computer Vision}'s Action Recognition Challenge, ICCV 2021.

% Without using any pre-trained weights, the proposed solution achieves a Top-1 accuracy of \textbf{73\%} on Kinetics400ViPriors test set for Data-Efficient Deep Learning Workshop’s Action Recognition Challenge, ECCV 2020, which is the best among all of the other entries.  

\end{abstract}

\section{Introduction}

Deep learning has enabled progress in video understanding tasks like action recognition~\cite{feichtenhofer2020x3d,kenshohara,demir2021tinyvirat}, action detection~\cite{rizve2021gabriella,duarte2018videocapsulenet,trecvid20,Rana_2021_WACV} temporal action localization \cite{Tirupattur_2021_CVPR,swetha2021unsupervised} etc. Most of the advancement in the video understanding due to deep network is built upon existence of the large scale labeled data like Kinetics~\cite{kinetics}, HACS~\cite{hacs}, LSHVU~\cite{diba2020large} etc.

% 1 paragraph citing recent video self-supervised papers
% 3D-CNNs are notorious to optimize with a small number of video samples and 

Recent works in video self-supervised learning~\cite{tclr,pace_pred,vcp,csj,cvrl,simon} show that spatio-temporal representations learned from the self-supervised learning on the same dataset also helps in improving results of the video encoder by a significant margin over the training from scratch. In our proposed solution we opt for the Temporal Contrastive Learning for video Representation (TCLR) method \cite{tclr} which gets the maximum gain among all methods while pretraining from the same dataset. On UCF101, without using any additional labeled or unlabeled data, TCLR pretrained model results in a boost of \textbf{20\%} Top-1 accuracy over the baseline model. 
% TCLR gets \textbf{20\%} boost on UCF101 over the baseline model without using any additional labeled or unlabeled data.

% \textcolor{brown}{Naman: write 1-2 lines praising vision transformers in general, then put video transformer stuff here, then MViT}

Recently, transformers have been applied to key computer vision tasks such as image classification after the introduction of  Vision Transformer (ViT)~\cite{dosovitskiy2021image}. The impressive performance of transformers in the image domain led to investigation of Transformer-based architectures for video-based classification tasks. Video transformers have lead to state of the art performances on Kinetics-400 \cite{carreira2017quo}, SSv2 \cite{goyal2017something} and Charades \cite{sigurdsson2016hollywood}. Adding temporal attention encoder on top of ViT~\cite{dosovitskiy2021image}(Pretrained) was proposed in VTN~\cite{neimark2021video} which led to good performance on video action recognition. A factorized spacetime attention based approach was proposed in TimeSformer~\cite{bertasius2021spacetime} after analysis of various variants of space-time attention based on compute-accuracy tradeoff. Video Swin Transformer~\cite{liu2021video} investigated spatiotemporal locality and showed that an inductive bias of locality a better speed-accuracy trade-off compared to other approaches which use global self-attention.
% MViT~\cite{fan2021multiscale} are transformers for video and image recognition which are trained from scratch and they  
In our proposed solution we adopt MViT~\cite{fan2021multiscale} transformers which shows state-of-the-art performance on Kinetics-400 without requiring any pretraining checkpoint unlike other video transformers architectures which requires ImageNet~\cite{deng2009imagenet} pretraining. Apart from eliminating the pretraining requirement, another advantage of using MViT is low computational requirement due to its pooling attention for spatiotemporal modeling.
% We have trained a variant of MViT\cite{fan2021multiscale} on \textit{Kinetics400ViPriors}.

%  By only having a first
% layer that ‘patchifies’ the input in spirit of a 2D convolu-
% tion, followed by a stack of transformer blocks, the vision
% transformer aims to showcase the power of the transformer
% architecture using little inductive bias.

% Action Recognition requires spatio-temporal understanding of a sequence of video frames. In literature, there are many works that tackle the action recognition problem by deep networks by learning spatio-temporal representation using convolutional operations on the RGB frames and/or motion priors such as optical flow \cite{zach2007duality}, dense point trajectories \cite{wang2013action}). The main approaches for action recognition are based on 2D ConvNets with LSTM \cite{yue2015beyond}, Two-stream architectures \cite{carreira2017quo}, \cite{simonyan2014two}, and single RGB stream 3D ConvNets \cite{tran2018closer}, \cite{tran2015learning}. Most of these approaches achieve high results on smaller datasets like UCF101 \cite{soomro2012ucf101} and HMDB \cite{jhuang2011large}  and Kinetics400ViPriors by using pretrained weights from ImageNet \cite{deng2009imagenet}, Sports-1M \cite{karpathy2014large}, or Kinetics-400 \cite{carreira2017quo}.

While learning from scratch, it is difficult to optimize the parameters of a 3D ConvNet based architecture with a single stream of RGB video frames from relatively smaller datasets such as Kinetics400ViPriors as compared to Kinetics-400 \cite{carreira2017quo} as the given dataset has same number of classes as Kinetics400 but roughly $\sim$20\%  of the number of videos in  Kinetics-400. Carreira et al. \cite{carreira2017quo} show that two-stream-based 3D ConvNet approaches significantly surpass single stream RGB video-based 3D ConvNet approaches; there is a $\sim$30\% improvement for the task of action recognition on both UCF101 and HMDB when no pre-trained weights are used. Also, \cite{davekallis} shows that optical flow is a powerful prior for modeling motion information while learning from scratch.

% Our approach is a two-stream based architecture, incorporating various spatio-temporal resolution clips for both RGB and optical flow streams. While testing a single video, we take the average of the predictions over multiple clips of that same video, which consists of various spatio-temporal resolutions as well as some simple augmentations, such as a horizontal flip. We show that this testing strategy boosts the performance of the action classifier, while also reducing the training cost significantly. 
We use an ensemble of various TCLR self-supervised pretrained 3D ConvNets and video transformers in the RGB stream and an ensemble of various 3D ConvNets in the optical flow streams. %We use an ensemble of various 3D ConvNet and MViT\cite{fan2021multiscale} in both RGB and optical flow streams which 
This helps in mitigating common generalization errors as well as decreasing the variance in neural network predictions.

% % MVIT
% A shift in backbone architectures for computer vision, from CNNs to Transformers, began recently with Vision Transformer (ViT)\cite{dosovitskiy2021image}. The great success of image Transformers has led to investigation of Transformer-based architectures
% for video-based classification tasks. VTN\cite{neimark2021video} proposes to add a temporal attention
% encoder on top of the pre-trained ViT, which yields good performance on video action recognition.
% TimeSformer\cite{bertasius2021spacetime} studies five different variants of space-time attention and suggests a factorized spacetime attention for its strong speed-accuracy tradeoff. MViT\cite{fan2021multiscale} is a multi-scale
% vision transformer for video recognition trained from scratch that reduces computation by pooling
% attention for spatiotemporal modeling, which leads to state-of-the-art results on SSv2.

\section{Proposed Method}

\subsection{Self-supervised pretraining- TCLR}

TCLR self-supervised framework explicitly encourages the learning of temporally distinct video representations. TCLR framework consists of mainly three components:

\subsubsection{Instance Contrastive Loss}
In a mini-batch of $N$ different video instances, 2 clips are taken from a video instance and stochastically transformed using various geometric and appearance based transforms. Following the instance discrimation objective the 2 differently augmented clips are brought together in the representation space whereas the clips from the different instances are pushed further apart using Instance Contrastive Loss ($\mathcal{L}_{IC}$) 

% \vspace{-5mm}
% \begin{align}\label{eq:IC}
% \small
%   \mathcal{L}_{IC}^{i}=-\log \frac{h\left(G_{i}, G'_{i}\right)}{\sum_{j=1}^{N}[\mathbb{1}_{[j\neq i]} h(G_{i}, G_{j}) + h(G_{i}, G'_{j})]}.
% \end{align}

\begin{equation}\label{eq:IC}
\small
  \mathcal{L}_{IC}^{i}=-\log \frac{h\left(G_{i}, G'_{i}\right)}{\sum_{j=1}^{N}[\mathbb{1}_{[j\neq i]} h(G_{i}, G_{j}) + h(G_{i}, G'_{j})]},
\end{equation}

\noindent where, $(G_{i}, G'_{i})$ are two clip representations from same instance $i$,  $h(u, v)=\exp \left(u^{T}v/(\|u\| \|v\| \tau) \right)$ is used to compute the similarity between $u$ and $v$ vectors with an adjustable parameter temperature, $\tau$. $\mathbb{1}_{[j\neq i]} \in \{0, 1\}$ is an indicator function which equals 1 iff $j \neq i$.

\subsubsection{Local-Local Temporal Contrastive Loss}
For this loss, we treat non-overlapping clips sampled from different temporal segments of the same video instance as negative pairs, and randomly transformed versions of the same clip as a positive pair. This allows the model to learn differences between timestamps of a video. The loss is defined as 

\begin{equation}
\label{eq:ccont}
\small
    \mathcal{L}_{LL}^{i}=-\sum_{p=1}^{N_T}\log\frac{h\left(G_{i,p} , G'_{i,p}\right)}{\sum_{q=1}^{N_{T}}[\mathbb{1}_{[q\neq p]} h(G_{i,p} , G_{i,q})+h(G_{i,p} , G'_{i,q})]}.
\end{equation}

where, $G_{i,p}$ represents a clip from instance $i$ at time $p$ and $G'_{i,p}$ represents the transformed version of the same clip. The positive pairs for this loss are formed by two clips from the same instance $i$ and the same timestamp $p$ (e.g. $G_{i,p}$ and $G'_{i,p}$). Any two clips from different timestamps of an instance $i$ are treated as a negative pairs (e.g. $G_{i,p}$ and $G_{i,q}$ form a negative pair). A given video instance $i$ is divided into $N_{T}$ non-overlapping clips. Hence, for every positive pair, the local-local contrastive loss has $2 \times N_{T} - 2$ negative pairs.

\subsubsection{Global-Local Temporal Contrastive Loss}
The purpose behind global-local temporal contrastive loss  is  to encourage the model to learn features that represent the temporal locality of the input clip across the temporal dimension of the feature map.

\begin{equation}\label{eq:tcp}
 \small
    \mathcal{L}_{GL_{k}}^{i}= \log\frac{h\left(L_{i,k} , G_{i,k}\right)}{\sum_{q=1}^{N_{T}} h(L_{i,k}, G_{i,q})} + \log\frac{h\left(G_{i,k}, L_{i,k}\right)}{\sum_{q=1}^{N_{T}} h(G_{i,k}, L_{i,q})}, 
\end{equation}
% \vspace{-5mm}
\begin{equation}\label{eq:tcp1}
 \small
    \mathcal{L}_{GL}^{i}= -\sum_{k=1}^{N_T}\mathcal{L}_{GL_{k}}^{i}.
\end{equation}

Where, $L_{i,k}$ represents the local clip from instance $i$ for timestamp $k$, and $G_{i,k}$ represents the features that are pooled from a global clip of $i$ but represent the same timestamp $k$. Hence, there are two separate ways to represent the timestamp $k$ of any instance $i$. This loss has two sets of reciprocal terms, with $G_{i,k}$ and $L_{i,k}$ serving as the anchor for each term and creating a positive pair. The negative pairs are supplied by matching the anchors with representations corresponding to other non-overlapping local clips. Note that similar to our local-local temporal contrastive loss we do not use negatives from other video instances for calculating this loss.%in Equation~\ref{eq:tcp}. 

% The schematic diagram of the proposed method is depicted in Fig 1. We use a two-stream based architecture. This architecture further consists of an ensemble of either different backbone architectures or the same architecture trained with different spatio-temporal resolutions.
% \begin{figure}[h]
% \begin{center}
% \includegraphics[width= \linewidth]{images/3.pdf}
% \end{center}
% \caption{Schematic Diagram of the proposed method (inference mode)}
% \label{fig:loss2}
% \end{figure}

% For the RGB stream, we use an ensemble of Inflated InceptionV3 (I3D) \cite{carreira2017quo}, C3D \cite{tran2015learning}, R2+1D -18 layer \cite{tran2018closer}, , ResNet 3D -18 layer and MViT\cite{fan2021multiscale} architectures with the same input. The three major types of augmentations that are applied to the RGB frames are, (1) Spatial augmentations: random crop, random scaling, and horizontal flip (2) Appearance transformation: random grayscale and color jittering (3) Temporal Augmentation: an evenly spaced, random number of skip frames and random starting frame. Each model is trained individually with all of the above augmentations. While testing, spatial and temporal augmentations are utilized. The augmentations aid in preventing overfitting while training on a small dataset such as UCF101. They also aid in mitigating the variance in the inference by taking the average of the predictions for the varying augmented clips within the same video.  
\subsection{Multiscale Vision Transformer}

% \textcolor{brown}{write only working of the MViT here, put some equation or diagram}

\begin{figure}[t!]
\begin{center}
% \fbox{\rule{0pt}{2in} \rule{.9\linewidth}{0pt}}
\includegraphics[width=0.55\linewidth]{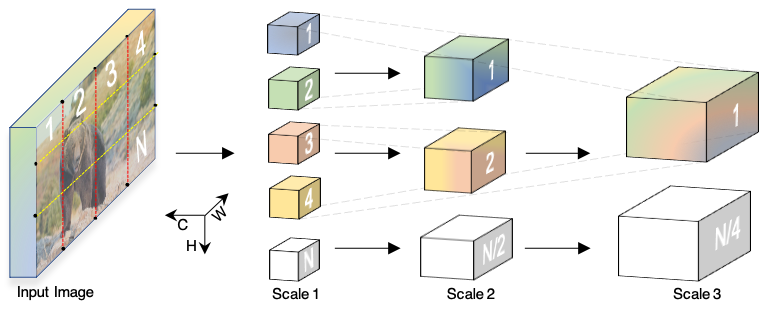}
\includegraphics[width=0.43\linewidth]{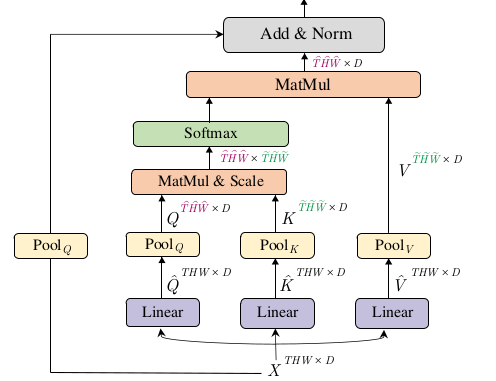}
\end{center}
  \caption{The left picture shows an overview of MViT's approach to learn from dense simple features to courser and complex features as the number of channels increase and resolution decreases. The right picture shows an overview of the flexible Pooling Attention mechanism of MViT\cite{fan2021multiscale}}
\label{mvit}
\end{figure}

% A shift in backbone architectures for computer vision, from CNNs to Transformers, began recently with Vision Transformer (ViT)\cite{dosovitskiy2021image}. The great success of image Transformers has led to investigation of Transformer-based architectures
% for video-based classification tasks. 
% VTN\cite{neimark2021video} proposes to add a temporal attention
% encoder on top of the pre-trained ViT, which yields good performance on video action recognition.
% TimeSformer\cite{bertasius2021spacetime} studies five different variants of space-time attention and suggests a factorized spacetime attention for its strong speed-accuracy tradeoff. 
MViT\cite{fan2021multiscale} is a multi-scale
vision transformer %for video recognition
which is trained from scratch.
% they use a pooling attention for spatiotemporal modeling approach for reduction in computation.
% The central advance of MViT~\cite{fan2021multiscale} is developing a spatiotemporal feature hierarchy within the Transformer backbone. 
Contrary to the typical Vision Transformer~\cite{dosovitskiy2021image} models which use a constant feature dimension and resolution in all layers and an attention mechanism to determine which previous tokens to focus, MViT~\cite{fan2021multiscale} proposes a flexible Multi Head Pooling Attention mechanism that pools the projected query, key, and value vectors, enabling reduction of the visual resolution. The right of Figure~\ref{mvit} gives overview of Multi Head Pooling Attention approach proposed by MViT~\cite{fan2021multiscale}.
 
This pooling attention is combined with an increase in the channel dimension with the idea of hierarchical feature construction from simple features which have high visual resolution and lower dimensions to more complex features having higher dimension features with lower resolution. The left of Figure~\ref{mvit} gives overview of hierarchical feature construction approach of MViT.

% We train MViT model taking 16 frames with frame rate of 4 on the Kinetics400ViPriors dataset.
% \subsection{CRF for Sequence Labeling}

\subsection{Optical Flow}
Optical flow has shown to increase performance on several video related tasks. In our experiments we calculate TVL-1 optical flow \cite{zach2007duality} for the dataset and use those features for training. The TVL-1 algorithm is based on a minimization of the energy function:

\begin{equation}\label{eq:tcp}
 E(\textbf{u}) = \int_{\Omega} |\nabla{u_{1}}| + |\nabla{u_{2}}| + \lambda|\rho(\textbf{u})|
\end{equation}
% For the TV-L1 stream, we use different spatio-temporal resolutions with two I3D models. The first model (I3D-flow-1) is trained with a stacked optical flow of 64 consecutive frames, with random cropping and horizontal flipping as augmentations. This model is expected to learn the fine-grained temporal structure of an activity. The second I3D model (I3D-flow-2) is trained with a stacked optical flow of 16 frames with evenly spaced, dynamic skip frames. This model sees the activity at a different temporal resolution, which helps in learning the coarse temporal features. For dynamic skip frame, a random number of temporal stride from 1 to $n_{max}$ is chosen, where, $n_{max}\,= \, floor(video \,frame \, count/ \,clip \,frame \, count)$ i.e. maximum skip frames possible in a clip. More details on the augmentations and resolutions are given in Table \ref{table1}, 
% \usepackage{vcell}
% \usepackage{rotating}

\section{Experiments}
This section covers details of the challenge dataset, implementation details, and results.  

\subsection{Dataset}

We use the provided Kinetics400ViPriors dataset, a modification of the official Kinetics400\cite{kay2017kinetics} dataset. The challenge dataset consists of 40k videos for training, 10k videos for validation, and 20k videos for testing. We use TV-L1 optical flow computed by the \cite{zach2007duality} method.  

\subsection{Implementation Details}
We perform TCLR self-supervised pretraining on train+val set without using any labels. The pretraining is performed for 400 epochs for R3D-18 and R3D-50 architectures following the learning rate schedule of \cite{tclr}. After the pretraining, each model is finetuned for the 150 epochs. More details input clip during training and inference setting is given in Table~\ref{table:resulttable}.

We have used the MViT-B with a convolutional pooling function as described in \cite{fan2021multiscale} which consists of 4 scale stages. The model takes 16 frames of 224$\times$224 resolution as input with a skip rate of 4. We used the code provided in MViT~\cite{fan2021multiscale} paper\footnote{\url{https://github.com/facebookresearch/SlowFast}}. We have followed truncated normal distribution initialization~\cite{hanin2018start} and trained for 200 epochs with 2 repeated augmentation~\cite{Hoffer_2020_CVPR} repetitions as described in MViT~\cite{fan2021multiscale}.

\subsection{Results}
% \textcolor{red}{needs to be written}

% We evaluate our trained model on multiple clips from the same video. More details on the training and testing methods for each model are provided in Table .%\ref{table1}.

We performed our initial experiments on Kinetics400ViPriors validation set by training on just training set to observe the performance of our ensemble and model selection purposes, shown in Table~\ref{table:resulttable}. The best models from the validation set performance are later finetuned on the train+val set for 100 epochs, and submitted for the test set evaluation. The proposed method achieves a Top-1 accuracy of 73\% on Kinetics400ViPriors test set without using additional data in our training. 
%We performed our training and model selection on the VI Priors Action Recognition dataset in the same manner.
% The final fused model (row-9) of Table-1 is used to predict the output of the competition test set and achieves a Top-1 accuracy of 90.71\%.  

% \textcolor{green}{rohit: fix the table}
\begin{table}

\centering
\small
\caption{Overview of accuracies of approaches used on val set}
\begin{tabular}{llcccc} 
\hline

\hline

\hline\\[-3mm]
 \# & Model       & Resolution & \begin{tabular}[c]{@{}l@{}}Frames \\$\times$ skip\end{tabular} & \begin{tabular}[c]{@{}l@{}}Test Time Crops\\(spatial $\times$ temporal)\end{tabular} & \begin{tabular}[c]{@{}l@{}}Validation\\Accuracy\end{tabular}  \\ 
\hline

\hline

\hline\\[-3mm]
1     & MViT        & 224        & 16 x 4                                                              & 3 x 10                                                                                & 51.3\%                                                       \\ 
% \hline
2     & I3D-OF      & 224        & 16 x 2                                                              & 3 x 10                                                                                & 40.5\%                                                       \\ 
% \hline
3     & I3D-OF      & 224        & 16 x 4                                                     & 3 x 5                                                                                 & 42.1\%                                                       \\ 
% \hline
4     & R3D18-OF    & 112        & 16 x 2                                                              & 3 x 10                                                                                & 31.9\%                                                       \\ 
% \hline
6  & R3D-50 TCLR &    224        & 16 x 2                                                              & 3 x 10                                                                                & 61.1\%                                                       \\ 
% \hline
6     & R3D-18      & 112        & 16 x 2                                                              & 3 x 10                                                                                & 33.2\%                                                       \\ 
% \hline
7     & R3D-18 TCLR & 112        & 16 x 2                                                              & 3 x 10                                                                                & 46.0\%                                                       \\
\hline
\hline
\label{table:resulttable}
\end{tabular}
\end{table}

\section{Conclusion}%\textcolor{red}{needs to be written}
In this technical report, we have shown that the self-supervised pretraining improves results significantly while learning from limited data without pretraining from additional labeled or unlabeled data. Combining the self-supervised 3D-CNNs with the state-of-the-art video transformer models and optical flow performs competitively on the test set. We believe our method can be further improved by pretraining the video transformer models in self-supervised manner on the same dataset.

% we presented a two-stream architecture based action classifier, using various spatio-temporal resolutions and augmentations in both training and inference. We use an ensemble of different architectures with the same input as well as an ensemble of the same architecture with inputs of different spatio-temporal resolutions. We also observed that taking an average of the predictions over multiple augmented clips aids in boosting the action recognition performance as well as reducing computational cost in training. 

%
\bibliographystyle{splncs04}
\bibliography{main}
\end{document}